# Exponential Natural Particle Filter

**Ghazal Zand, Mojtaba Taherkhani and Reza Safabakhsh**

Robotics Research Institute, AmirKabir University of Technology, Tehran, Iran
Computer Engineering and Information Technology Dep., AmirKabir University of Technology, Tehran, Iran
{ ghazal.zand, taherkhani.mojtaba, safa }@aut.ac.ir

## Abstract

Particle Filter (PF) is the most widely used Bayesian sequential estimation method for obtaining hidden states of nonlinear dynamic systems. However, it still suffers from certain problems such as the loss of particle diversity, the need for large number of particles, and the costly selection of the importance density functions. In this paper, a novel PF called Exponential Natural Particle Filter (xNPF) is introduced to solve the above problems. In this approach, a state transitional probability with the use of natural gradient learning is proposed which balances exploration and exploitation more robustly. PF with the proposed density function does not need a large number of particles and it retains particles' diversity in a course of run. The proposed system is evaluated in a time-varying parameter estimation problem on a dynamic model of HIV virus immune response. This model is used to show the performance of the xNPF in comparison with several state of the art particle filter variants such as Annealed PF, Bootstrap PF, iterative PF, equivalent weight PF, and intelligent PF. The results show that xNPF converges much closer to the true target states than the other methods.

## Introduction

The states of a system provide a complete representation of the system's internal condition and status (Yin and Zhu 2015); hence obtaining the system states is very important for many problems such as process monitoring (Gao and Ding 2007; Gao and Ho 2006), localization (Lu 2014; Yang 2014), robot navigation (Atia et al. 2010; Hiremath et al. 2014), single or multi object tracking (Bae and Yoon 2014; Fan, Ji, and Zhang 2015; Mihaylova 2014; Nannuru and Coates 2013; Zhang 2015; Zhao 2013), lane detection (Shin, Tao, and Klette 2014) and so on. These problems require estimation of the state of a dynamical system by using a sequence of noisy measurements made on the system. Solving these problems is the goal of state estimation algorithms. Bayesian sequential estimation is the basis for the most of the state estimation methods. In a Bayesian framework, the aim is thus to compute the posterior distribution $p(X_t|Z_{1:t})$, where $Z_{1:t}$ is the noisy observation set from time 1 to t and $X_t$ is the global state vector at time t.

The Particle Filter algorithm (PF) is a Bayesian sequential estimation method whose fundamental idea is to approximate $p(X_t|Z_{1:t})$ with a set of particles with associated weights. The PF is designed for a hidden Markov Model, in which the observation processes are related to the state processes, as shown in Figure 1.

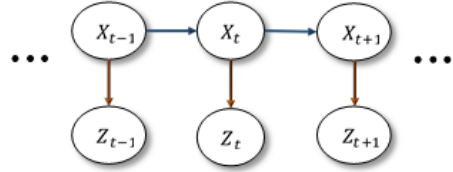

*Figure 1: Particle Filter's hidden Markov Model. This model shows how the observation processes are related to the state processes.*

$p(X_t|X_{t-1})$ is the state transition probability and $p(Z_t|X_t)$ is the observation probability by given state.

To briefly review generic PFs, let's consider the General form of discrete-time nonlinear dynamical systems as

$$X_{t+1}=f(X_t,Q_t) \qquad (1)$$
$$Z_t=h(X_t,E_t)$$

where, $Q_t$ and $E_t$ denote process noise and observation noise at time $t$, $f$ and $h$ are nonlinear functions of the system. The noise signals are assumed to be independent and with known Probability Density Function (PDF). The posterior density is calculated as

$$p(X_t|Z_{1:t})=k.p(Z_t|X_t)\int p(X_t|X_{t-1}).p(X_{t-1}|Z_{1:t-1})dX_{k-1} \qquad (2)$$

The integral in (2) can be approximated with $N$ independent weighted particles by using Monte-Carlo sampling. It starts with assuming that there is a weighted sample representation for the posterior $p(X_{t-1}|Z_{1:t-1})$ at the previous time step.

$$S_{t-1}=\left\{(x_{t-1}^{(i)}, w_{t-1}^{(i)}); 1,...,N\right\} \qquad (3)$$

where, $w_{t-1}^{(i)}$ and $x_{t-1}^{(i)}$ denote the weight and the state of particle $i$ at time $t-1$. By using this set to carry out Monte-Carlo integration, the integral in (2) is approximated as:

$$\int p(X_t|X_{t-1}).p(X_{t-1}|Z_{1:t-1})dX_{k-1} \approx \sum_i w_{t-1}^{(i)} p(X_t|x_{t-1}^{(i)}) \quad (4)$$

Then, the posterior density is calculated by re-weighting the particles as:

$$w_t^{(i)} \propto \frac{p(z_t|x_t^{(i)})p(x_t^{(i)}|x_{t-1}^{(i)})}{q(x_t^{(i)}|x_{t-1}^{(i)},Z_{1:t})} w_{t-1}^{(i)} \quad (5)$$

Where $q(x_t^{(i)}/x_{t-1}^{(i)},Z_t)$ is a proposal density function to estimate the particles' positions. The algorithms mostly use $p(x_t^{(i)}|x_{t-1}^{(i)})$ as the proposal density function.

After computing the weights, particles need to be resampled independently $N$ times, with replacement, from the obtained distribution. Resampling is performed to avoid degeneracy in the particle filter. In the resampling phase, particles with low importance are removed and replaced with more probable new particles. The diagram of Figure 2 (Smith et al. 2013) shows the mentioned iteration of the PF.

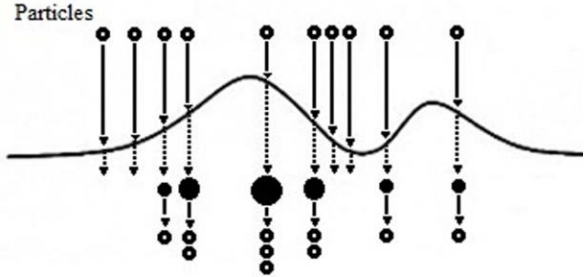

*Figure 2: Resampling phase in PF algorithm. In this phase, particles with low weight are replaced with more important particles.(Smith et al. 2013)*

The PF algorithm performs relatively better than other nonlinear filtering methods; but it still face some problems: (a) the importance density functions selection (Fan et al. 2015; Zuo et al. 2013). (b) the loss of particle diversity (Fan et al. 2015). (c) the need for large number of particles (Ades and Van Leeuwen 2013).

The objective of this paper is to design a new PF which tries to solve the above mentioned problems. In the proposed method, called Exponential Natural Particle Filter (xNPF), particles are classified into two classes. The particles of the first class aim to retain the diversity (as the Exploration stage) by using original state transition probability; whereas the particles of the second class try to search locally (as the Exploitation stage) by using the proposed state transition probability. In addition, the proposed PF does not need a large number of particles. The reason is that unlike the standard PF, in which a few of particles have a significant weight and the rest of them have a negligible weigh, in the proposed method the majority of particles contribute significant information to the posterior probability.

The rest of this paper is organized as follows. The second section provides a review of different PFs. In this section, PF algorithms are classified based on addressing the mentioned problems. In the third section, a novel particle filter is proposed and explained in detail. The experimental results are discussed in the fourth section, and the last section concludes the paper.

## Related Works

Since the introduction of PF(N. J. Gordon, Salmond, and Smith 1993), many intelligent search and optimization methods have been incorporated into PF to enhance the performance of the classic PF. For example, Bin et al. (Bin 2013) proposed a localization algorithm called PSO-PF algorithm based on particle swarm optimization. By propagating the particles using the PSO method, their algorithm eliminates the phenomenon of loss of particles' diversity.

In (Szczuko 2014), a Hybrid genetic-APF method is presented in which a genetic algorithm is combined with annealed PF for human body pose estimation. In these algorithms, particles share more information that causes locating in the most probable areas. In (Park et al. 2009; Yin and Zhu 2015), the genetic operators are designed to further improve the particle diversity while avoiding the premature convergence, since the misleading state estimating problem in general PF is mainly caused by the poverty of particle diversity.

Kalami and Khaloozadeh tried to solve the misleading state estimating problem by converting the nonlinear state estimation problem to a multi-objective optimization problem and proposed non-dominated Genetic Filter based on the genetic algorithm and PF (Kalami Heris and Khaloozadeh 2014).

Recently, Fan et al. proposed an Iterative PF (IPF) which converges to true target state more closely by sampling the particles iteratively with the search scope contracted (Fan et al. 2015). IPF is motivated by the success of both increasing the sampling density and simulated annealing.

As mentioned before, the limitation of the number of particles that can be used, is one of the most serious issues in the particle filtering problem. To overcome this problem, BoxPF (Gning, Ristic, and Mihaylova 2012; Gning et al. 2013) addresses the concept of box particle that can make sampling from coarse to fine, improving the sampling efficiency. The purpose of this method is to reduce the number of random samples required by the standard particle filter. Ades et al. proposed a PF which uses the proposal density to ensure that all particles end up in the high-probability region of the posterior pdf (Ades and Van Leeuwen 2013). This gives rise to the possibility of nonlinear data assimilation in large-dimensional systems.

Also, in (Ades and Van Leeuwen 2015), a fully nonlinear equivalent weights PF is presented to avoid the need for a large numbers of particles by replacing the standard model transition density with two different proposal transition densities. The first one is used to relax all particles towards the high-probability regions of the state space as defined by the observations. The second one is then used to ensure that the majority of particles have equivalent weights at the observation time.

Yoon et al. describe a vision-based system for tracking objects based on fuzzy PF algorithm in which the states are estimated using a fuzzy expected value operator (Yoon, Cheon, and Park 2013). This fuzzy PF algorithm is proposed to overcome problems from the occurrence of the unexpected improper variances due to several causes (e.g. process noises, or observation noises).

Finally, in the last decades many famous modifications have been introduced in the PF, such as the Bootstrap Particle Filter (BPF), the Auxiliary Particle Filter (APF), the Regularized Particle Filter (RPF), the Unscented Particle Filter (UPF), and the Gaussian Particle Filter (GPF)(Arulampalam et al. 2002; N. Gordon, Ristic, and Arulampalam 2004; N. J. Gordon et al. 1993; Kotecha and Djurić 2003).

## Exponential Natural Particle Filter

The PF algorithm proposed in this paper, which is called the Exponential Natural Particle Filter (xNPF), is inspired by the following idea.

The appropriate selection of the proposal density function can increase the sampling density to close to the true state, so that the particles can be used sufficiently. Thus, it can meet the same precision requirements with fewer particles. On the other hand, increasing the sampling density in a small region leads to the loss of diversity. So, the particles cannot search the environment properly. In the xNPF, particles are classified into two classes A and B. The particles of class A try to retain the diversity (exploration) by using original state transition probability, and the particles of class B try to search locally (exploitation) by using the proposed state transition probability. To classify particles, a partitioning parameter ($\pi_{partition}$) $\in [0,1]$ is considered.

This parameter randomly partitions all particles into class A and class B, in which the cardinalities are defined with

$$|A|=\lceil \pi_{partition} * N \rceil \qquad (6)$$
$$|B|=N-|A|$$

where $\lceil x \rceil$ indicates the smallest integer which is greater than or equal to x (ceiling of x).

The idea of using natural gradient learning has been proposed in Exponential Natural Evolution Strategy (xNES). To propose the state transition probability for class B, xNES is used which iteratively updates the probability distribution by using an estimated gradient on its distribution parameters (Glasmachers et al. 2010). These updates continue until T iteration. This estimation of distribution algorithm presents an exponential parametrization of the search distribution. It guarantees invariance, while at the same time provides an elegant and efficient way of computing the natural gradient without the need for the explicit Fisher information.

In this paper by selecting a large value for the step size which is used in the original transitional probability, diversity of the particles of class A is retained while the exploration is improved. On the other hand, by using an estimated gradient, the proposed transitional probability for class B improves local search ability of the particles. Thus, the balance between global and local search is gained.

Using the above idea, the proposal density function for estimating the particles' positions is defined as

$$q(x_t^{(i)}) = \begin{cases} p(x_t^{(i)}|x_{t-1}^{(i)}) & i \epsilon A \\ q_{xNES}(x_t^{(i)}|x_{t-1}^{(i)}, m_{t-1}, \sigma_{t-1}^2, C_{t-1}) & i \epsilon B \end{cases} \qquad (7)$$

where $q_{xNES}(./.)$ is the proposed state transition probability which is computed by xNES; $m_{t-1}$, $\sigma_{t-1}$ and $C_{t-1}$ are the mean value, step size, and covariance matrix returned by xNES at the end of the $T^{th}$ iteration.

According to the (6), the primary probabilites of classes A and B are calculated as

$$Pr\{i \epsilon A\} = 1 - \pi_{partition} \qquad (8)$$
$$Pr\{i \epsilon B\} = \pi_{partition}$$

Hence, (7) can be written in compact form as

$$q(x_t^{(i)}|x_{t-1}^{(i)}, Z_t) = \pi_{partition} q_{xNES}(x_t^{(i)}|x_{t-1}^{(i)}, m_{t-1}, \sigma_{t-1}^2 C_{t-1}) \quad (9)$$
$$+ (1-\pi_{partition})p(x_t^{(i)}|x_{t-1}^{(i)})$$

According to (5) and (7), the rule of updating weights in xNPF is given by

$$w_t^{(i)} \propto \qquad (10)$$
$$\frac{p(z_t|x_t^{(i)}) \, p(x_t^{(i)}|x_{t-1}^{(i)})}{\pi_{partition} q_{xNES}(x_t^{(i)}|x_{t-1}^{(i)}, m_{t-1}, \sigma_{t-1}^2 C_{t-1}) + (1-\pi_{partition})p(x_t^{(i)}|x_{t-1}^{(i)})} w_{t-1}^{(i)}$$

The resampling method in the standard PF is known as the Roulette Wheel Selection (RW) rule in the field of evolutionary computation. Methods like RW may have a bad performance when a particle has a high weight in comparison with other particles. In RW, the most highly weighted particle may saturate the candidate space, and this leads to more loss of particles' diversity.

For the resampling step of xNPF, the Stochastic Universal Sampling (SUS) (De Falco et al. 2012) algorithm is used. This algorithm does not have the mentioned problem. SUS uses a single random value to sample all of the particles by choosing them at evenly spaced intervals. This gives weaker particles a chance to be chosen and thus reduces the unfair nature of fitness-proportional selection methods. The other parts of xNPF are the same as these in the standard particle filter. The overall structure of the xNPF algorithm is shown in Figure 3.

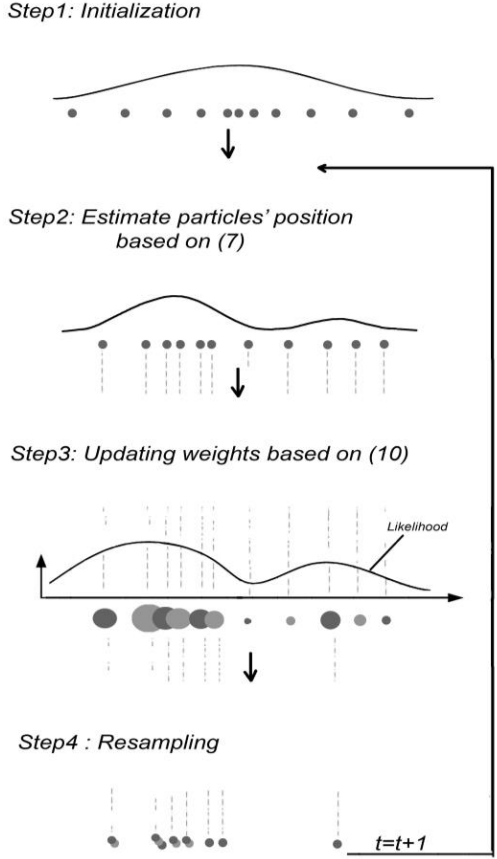

Figure 3: Overall structure of the xNPF algorithm. In each step, the concept of the process is schematically shown. The size of the circles represents the associated weight.

## Experiments

The efficiency of the xNPF is demonstrated by applying it to the problem of state estimation of the HIV infection model (Hartmann, Vinga, and Lemos 2012) assuming parametric uncertainty. To mathematically model the Human Immunodeficiency Virus (HIV), the three-dimensional basic infection model is considered. The model is given as

$$\dot{T} = s - dT - \beta(t)Tv$$
$$\dot{T}^* = \beta(t)Tv - \xi T^* \quad (11)$$
$$\dot{v} = kT^* - cv$$

where $T$ is the concentration of healthy CD4+ T-cells, $T^*$ is the concentration of HIV-infected CD4+ T-cells, and $v$ is the free virus particles. The six parameters can be written in a vector form as $\theta=(s, d, \beta, \xi, k, c)$, where all components have biological significance.

In a nutshell, (11) can be interpreted as follows. Healthy CD4+ T cells, which have an average life span of $1/d$ days, are produced at a constant rate $s$. These cells can get infected by free virus particles. The infection is modeled using a simple mass-action type term, with a time-varying parameter $\beta$. Infected cells may have a different life span $(1/\xi)$ than healthy cells, which means in general $\xi \neq d$. Finally, free virus particles are produced in infected cells, and released at a rate $k$, having an average life span of $1/c$. Figure 4 presents the mentioned HIV infection model.

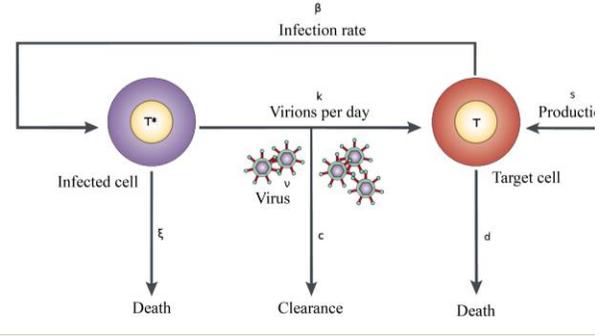

Figure 4: The dynamic model of HIV infection. Healthy T cells are infected by virus with rate $\beta$. Target T cells are assumed to be produced at rate $s$ and to die at rate $d$ per cell. Infected cells $T^*$, which produce new virions at rate $k$, and die at rate $\xi$ per cell (Perelson 2002).

In this paper, the numerical values for the parameters of model (11) are selected as follows (Hartmann et al. 2012): $\theta=(368.94, 0.46, 7.26e-06, 2.16, 1317.4, 3.6)$. Simulation time is set to 190 days and a 25 days periodic input is applied to the system which results in the parameter $\beta=3.63e-04$. The measurement equation related to system is given by

$$Z_t = \begin{bmatrix} 1 & 1 & 0 \\ 0 & 0 & 1 \end{bmatrix} \begin{bmatrix} T_t \\ T_t^* \\ v_t \end{bmatrix} + \begin{bmatrix} e_{1,t} \\ e_{2,t} \end{bmatrix} \quad (12)$$

where $e_{1,t} \sim N(0, \sigma_1^2)$ and $e_{2,t} \sim N(0, \sigma_2^2)$ are independent zero-mean Gaussian noises. In this paper, it is assumed that $\sigma_1^2 = 0.05$ and $\sigma_2^2 = 1$.

In the first subsection, the effect of varying the number of particles and the partitioning parameter will be analyzed. In the second subsection, the performance of the xNPF is compared with some particle filter variants. The performance measure used to investigate the effectiveness of strategies is the RMSE.

The default parameters for xNPF, such as the simulation time, the maximum number of xNES iterations, the number of particles, partitioning parameter and the step size of the original transitional probability are given in Table 1. The results of all experiments are averaged over 30 independent runs.

*Table 1: The default values for parameters of the xNPF.*

| Symbol | Description | Range | Default Value |
|---|---|---|---|
| iter | Maximum xNES iterations | $[1,\infty)$ | 5 |
| $\sigma 0$ | Step size of $p(x_t|x_{t-1})$ | $(0,\infty)$ | 10 |
| N | The number of particles | $[1,\infty)$ | 10 |
| $\pi_{partition}$ | Partitioning parameter | $[0,1]$ | 0.2 |

**The number of particles and partitioning parameter**

To investigate the effect of increasing the number of particles on the performance of the proposed method, various values of *N* are applied to the HIV infection model. For this experiment xNPF is run for different values of *N* and the default values of other parameters as in Table 1. Figure 5(a) shows the occurrence probability of the normalized weight versus the number of particles. Figure 5(b) and Figure 5(c) show the RMSE versus the number of particles.

The problem in the standard PF is that the majority of the particles are far from observation; thus only a few of them have a significant weight and the rest have negligible weights. Then to achieve a satisfactory performance, the PF algorithm turns out to require a massive number of particles, which induces high computational complexity. Figure 5(a) shows that with the proposed density function, the particles are more likely to get closer to the true states and get better weights. Thus the method avoids blind sampling, by which the majority of the particles contribute significant information to the probability of the system state given the observation.

By avoiding blind sampling, increasing the number of particles has no significant effect on bringing the approximation error down. Figure 5(b) and 5(c) show the mentioned result. Observing the graph closely, the obtained RMSE values of xNPF with the number of particles within the range of [25, 200] are similar. However, for comparing with other PF variants, *N* is considered equal to 25. Changes of the error value within the range of [5, 25] is quite natural since there is a small number of particles and xNPF cannot search the environment properly.

To study the effect of the partitioning parameter on the performance of xNPF, various configurations of the partitioning parameter are used. For this reason, xNPF is run for different values of $\pi_{partition}$ and the default values for other parameters as in Table 1. Performances of different configurations are summarized in Figure 6.

If the $\pi_{partition}$ assumes a small value, the number of particles in class B are increased. Thus the position of most of the particles is adjusted by gradient learning. To put it another way, these particles search locally (as the exploitation stage). Conversely, if the parameter assumes a large value, the number of particles in class A are increased. Thus the positions of the majority of the particles are adjusted in the same way as in the standard PF. These particles search globally (as the exploration stage) proportional to the value of the step size of $p(X_t|X_{t-1})$.

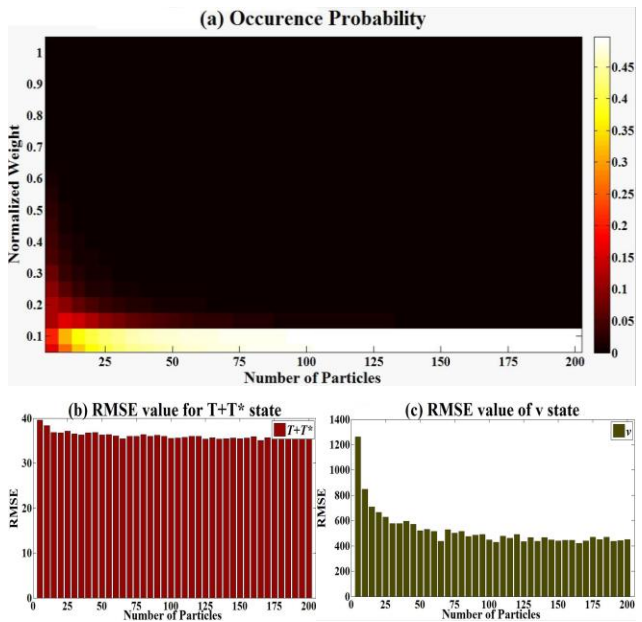

*Figure 5: the effect of different number of particles on performance of xNPF on HIV infection model (a) occurrence probability of normalized weight versus number of particles, (b) RMSE value for T+T\* state versus number of particles and (c) RMSE value for v state versus number of particles.*

To have a successful search, an optimal balance between the exploration and exploitation must be found. If the particles only exploit, they will not be able to cover the drastic changes in the sequent system states. As a result, it will be more likely to predict the system state unsuccessfully. In addition, when the observation model causes many local optima, the error of the system state prediction gradually increases. Gradient learning causes the particles get stuck in these local optima. If the particles only explore, they can follow the system state, but their prediction of the system state will not be very accurate. Thus the error values increase. Hence, some of the particles should be searching globally, so that they can follow the system states even in drastic changes. Furthermore, there should be other particles searching locally, so that they can predict the system state more accurately.

Figure 6(a) indicates that by increasing the value of $\pi_{partition}$, the changes in the prediction of state $T+T^*$ is slight. In other words, state $T+T^*$ is easily predictable during the simulation, but state *v* is not. The graph in Figure 6(b) shows that by increasing the value of $\pi_{partition}$, although the particles are able to follow the system state, the error increases because the particles are not able to predict the system state very accurately. In addition, since the sequent state *v* does not have drastic change, the small values for the $\pi_{partition}$ are good results. With the $\pi_{partition}$ within the range of [0.05, 0.3], error values are almost the same. However, to bring about a balance between the local and global search, $\pi_{partition}$ equal to 0.3 is considered. In such a situation, some of the particles search globally, whilst the rest of them make the prediction of the system states more accurately by gradient learning.

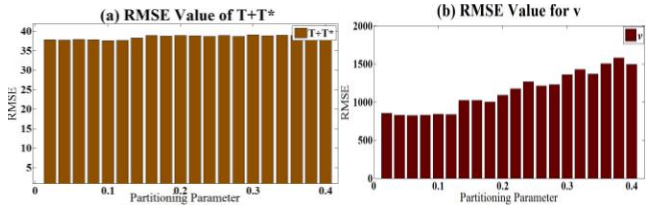

*Figure 6: partitioning parameter effect on performance of the xNPF the on HIV infection model (a) RMSE value for T+T\* state versus partitioning parameter (b) RMSE value for v state versus partitioning parameter*

**Comparison with other algorithms**

From all strategies introduced in the previous sections, five strategies are adopted in this study for comparisons: Annealed PF (APF)(Szczuko 2014), Bootstrap PF (BPF) (N. J. Gordon et al. 1993), iterative PF (IPF) (Fan et al. 2015), equivalent weight PF (Ades and Van Leeuwen 2015), and intelligent PF (Yin and Zhu 2015).

Based on our previous experiments, xNPF is run with $N$=25, $\pi_{partition}$=0.3 and the default values for other parameters as in Table 1. The other algorithms are run with $N$ equal to 100 and their other parameters are set based on the recommended values in their corresponding references.

For a better comparison of the results, Table 2 lists the mean and standard deviation of the RMSE and the number of particles' re-weighting found by each method. The bold number indicates the best solution according to a t-test with a significance level of 5%. Considering the same number of particles for all of the selected methods is not good for comparing the results properly. The methods may have some heuristic operators themselves, by which $p(X_t/X_{t-1})$ is calculated during several evaluations (going through re-weighting step). Thus, the number of re-weighing is used for comparing RMSE values.

Results indicate that all of the methods have the same RMSE in prediction of the state $T+T^*$. The RMSE values of IPF, Equivalent weight PF and xNPF were better than the other methods in the prediction of state $v$. But the re-weighing number of xNPF is much lesser than IPF and Equivalent weight PF. Thus, xNPF is generally more successful. This method uses two classes of particles for exploration and exploitation. As a result, it can follow the system states even in sequentially drastic changes and can converge to the system states more accurately.

*Table 2: State estimation comparison on HIV infection model*

|  | RMSE | | Number of re-weighing |
|---|---|---|---|
|  | T+T* | V |  |
| **BPF** | 38.221 (0.489) | 1.4379E+03 (20.587) | 190*100 |
| **APF** | 37.752 (1.026) | 807.800 (70.658) | ~190*500 |
| **IPF** | 37.100 (1.172) | 654.870 (48.255) | ~190*500 |
| **equivalent weight PF** | 38.094 (1.129) | 659.12 (51.801) | ~190*500 |
| **intelligent PF** | 38.491 (0.632) | 1.3140E+03 (18.910) | ~190*164 |
| **xNPF** | 37.128 (1.072) | 674.725 (50.620) | ~190*100 |

**Conclusions**

In this paper, the exponential natural particle filter has been introduced. With the proposed density function, this modification of the PF algorithm is able to efficiently converge to the true state. In this approach, a state transitional probability with the use of natural gradient learning is proposed which balances exploration and exploitation more robustly. PF with the proposed density function does not need a large number of particles and it retains particles' diversity in a course of run.

Experiments were designed in terms of comparing the solution accuracy and investigating the effect of increasing the number of particles and partitioning parameter on the performance of xNPF. The results clearly show the superiority of the xNPF over other novel methods. This method shows good experimental results due to using the proposed density function. It is possible to use an asymmetrical distribution as the proposal density function to consider the direction of the system states changes in the next states predictions.